\documentclass[runningheads]{llncs}

\usepackage[T1]{fontenc}
\usepackage{graphicx}
\usepackage{multirow}
\usepackage{todonotes}
\usepackage{xcolor}
\usepackage{comment}
\usepackage{adjustbox}
\usepackage{amsmath}
\usepackage[utf8]{inputenc}
\usepackage[english]{babel}
\usepackage{color}
\usepackage{amsfonts}
\usepackage{url}
\usepackage[mathscr]{euscript}
 \let\mathscr\relax
\usepackage[scr]{rsfso}
\usepackage{caption}
\usepackage[labelfont=bf]{caption}
\usepackage{tabularx,booktabs}
\usepackage{chngcntr}
\usepackage{newfloat}
\usepackage{multirow}
\usepackage{multicol}
\usepackage{adjustbox}
\DeclareFloatingEnvironment[name={Suuplementary Figure}]{suppfigure}

\DeclareFloatingEnvironment[name={Suuplementary Table}]{suppTable}

\usepackage{multicol}
\usepackage{algorithm2e}
\usepackage{booktabs,ragged2e}
\usepackage[utf8]{inputenc}
\usepackage{lineno}
\usepackage{makecell}

\usepackage[flushleft]{threeparttable}
\usepackage{subcaption}
\usepackage{color}

\author{
Ziang Xu  \inst{1,2}, Sharib Ali \inst{1,3}, Soumya Gupta \inst{1,2}, Simon Leedham\inst{4,5}, James E East \inst{4,5}, Jens Rittscher \inst{1,2,4}}
\authorrunning{Z. Xu, S. Ali et al.}
\institute{Institute of Biomedical Engineering, University of Oxford, Oxford, UK \and Big Data Institute, University of Oxford, Li Ka Shing Centre for Health Information and Discovery, Oxford, UK \and School of Computing, University of Leeds, Leeds, UK \and NIHR Oxford Biomedical Research Centre, Oxford, UK \and Translational Gastroenterology Unit, Experimental Medicine Div., John Radcliffe Hospital, University of Oxford, Oxford, UK\\}
\begin{document}
\title{Patch-level instance-group discrimination with pretext-invariant learning for colitis scoring}
\titlerunning{PLD-PIRL for colitis scoring}
\maketitle      
\begin{abstract}
Inflammatory bowel disease (IBD), in particular ulcerative colitis (UC), is graded by endoscopists and this assessment is the basis for risk stratification and therapy monitoring. Presently, endoscopic characterisation is largely operator dependant leading to sometimes undesirable clinical outcomes for patients with IBD. We focus on the Mayo Endoscopic Scoring (MES) system which is widely used but requires the reliable identification of subtle changes in mucosal inflammation. Most existing deep learning classification methods cannot detect these fine-grained changes which make UC grading such a challenging task. In this work, we introduce a novel patch-level instance-group discrimination with pretext-invariant representation learning (PLD-PIRL) for self-supervised learning (SSL). Our experiments demonstrate both improved accuracy and robustness compared to the baseline supervised network and several state-of-the-art SSL methods. Compared to the baseline (ResNet50) supervised classification our proposed PLD-PIRL obtained an improvement of 4.75\% on hold-out test data and 6.64\% on unseen center test data for top-1 accuracy.
\end{abstract}
\section{Introduction}
Ulcerative colitis (UC) is a chronic intestinal inflammatory disease in which lesions such as inflammation and ulcers are mainly located in the colon and rectum. UC is more common in early adulthood, the disease lasts for a long time, and the possibility of further cancerous transformation is high~\cite{torres2021results}. It is therefore important to diagnose ulcerative colitis early. Colonoscopy is a gold standard clinical procedure widely used for early screening of disease. Among the various colonoscopic evaluation methods proposed, the Mayo Endoscopic Score (MES) is considered to be the most widely used evaluation indicators to measure the UC activity~\cite{vashist2018endoscopic,d2007review}. MES divides UC into three categories, namely mild (MES-1), moderate (MES-2) and severe (MES-3). MES-2 and MES-3 indicate that an immediate follow up is required. However, the grading of UC in colonoscopy is dependent on the level of experience. Differences in assessment amongst endoscopists have been observed that can affect patient management. Automated systems based on artificial intelligence can help identify subtle abnormalities that represent UC, improve diagnostic quality and minimise subjectivity.

Deep learning models based on Convolutional Neural Networks (CNN)~\cite{lecun1998gradient} have already been used to build UC MES scoring systems~\cite{mokter2020classification,becker2021training}. But rather than formulating the problem as a 3-way classification task that separates the three MES categories (mild, moderate, severe), existing methods resort to learning binary classifiers to deal with the high degree of intra-class similarity. Consequently, a number of different models needs to be trained. 

We propose to amplify the classification accuracy of a CNN network for a 3-way classification using an invariant pretext representation learning technique in a self-supervised setting that exploits patch-based image transformations and additionally use these patches for instance-based group discrimination by grouping same class together using $k$-means clustering, referred to as ``PLD-PIRL''. The idea is to increase the intera-class separation and minimise the intra-class separation. The proposed technique uses a CNN model together with unsupervised k-means clustering for achieving this objective. The subtlety in the mucosal appearances are learnt by transforming images into a jigsaw puzzle and computing contrastive losses between feature embedding (\textit{aka} representations). In addition, we also explore the introduction of an attention mechanism in our classification network to further boost classification accuracy.
%
%
We would like to emphasise that the UC scoring is a complex classification task as image samples are very similar and often confusing to experts (especially between grades 1 and 2). Thus, developing an automated system for this task has tremendous benefit in clinical support system. 

The related work on UC scoring based on deep learning in presented in Section 2. In Section 3, we provide details of the proposed method. Section 4 consists of implementation details, dataset preparation and results. Finally, a conclusion is presented in Section 5.
\section{Related work}
\subsection{CNN-based UC grading}
Most research work on UC grading is based on MES scores. Mokter et al.~\cite{mokter2020classification} propose a method to classify UC severity in colonoscopy videos by detecting vascular (vein) patterns using three CNN networks and use a training dataset comprising of over 67k frames. The first CNN is used to discriminate between a high and low density of blood vessels. Subsequently they use two CNNs separately for the subsequent UC classification each in binary two class configuration. Such a stacked framework can minimise false positives but does not enhance the model's ability to understand variability of different MES scores. Similarly, Stidham \textit{\textit{et al.}}~\cite{stidham2019performance} use the Inception V3 model to train and evaluate MES scores in still endoscopic frames. They used 16k UC images and obtained an accuracy of 67.6\%, 64.3\% and 67.9\% for the three MES classes. UCS-CNN~\cite{alammari2017classification} includes several prepossessing steps such as interpretation of vascular patterns, patch extraction techniques and CNN-based classifier for classification. A total of 92000 frames were used for training obtaining accuracy of 53.9\% 62.4\% and 78.9\% for mild, moderate and severe classes.
Similarly, Ozawa \textit{et al.}~\cite{ozawa2019novel} use a CNN for binary classification only on still frames comprising of 26k training images, which first between normal (MES 0) and mucosal healing state (MES 1) while next between moderate (MES 2) and severe (MES 3). Gutierrez \textit{et al.}~\cite{becker2021training} used CNN model to predict a binary version of the MES scoring of UC.

One common limitation of existing CNN-based UC scoring literature is the use of existing multiple CNN models in an ensemble configuration, simplifying MES to a binary problem and use of very large in-house datasets for training. In contrast, we aim to develop a single CNN model-based approach for a 3-way MES scoring that is clinically relevant. In addition, we use a publicly available UC dataset~\cite{borgli2020hyperkvasir} to guarantee reproducibility of our approach. Futhermore, this dataset consists of only 851 image samples and therefore poses a small data problem. 
\subsection{Self-supervised approach for classification}
Self-supervised learning (SSL) uses pretext tasks to mine self-supervised information from large-scale unsupervised data, thereby learning valuable image representations for downstream tasks. By doing so, the limitation of network performance on predefined annotations are greatly reduced. In SSL, the pretext task typically applies a transformation to the input image and predicts properties of the transformation from the transformed image. Chen {\it \textit{et al.}}~\cite{chen2020simple} proposed the SimCLR model, which performs data enhancement on the input image to simulate the input from different perspectives of the image. Contrastive loss is then used to maximize the similarity of the same object under different data augmentations and minimised the similarity between similar objects. Later, the MoCo model proposed by He \textit{et al.}~\cite{he2020momentum} also used contrastive loss to compare the similarity between a query and the keys of a queue to learn feature representation. The authors used a dynamic memory, rather than static memory bank, to store feature vectors used in training. In contrast to these methods that encourages the construction of covariant image representations to the transformations, pretext-invariant representation learning (PIRL)~\cite{misra2020self} pushes the representations to be invariant under image transformations. PIRL computes high similarity to the image representations that are similar to the representation of the transformed versions and low similarity to representations for the different images. Jigsaw puzzle~\cite{noroozi2016unsupervised} was used as pretext task for PIRL representation learning.

Inspired by PIRL, we propose a novel approach that exploits the invariant representation learning together with patch-level instance-group discrimination. Here, the idea is to increase the inter-class separation and minimise the intra-class separation. An unsupervised $k$-means clustering is used to define feature clusters for $k$-class categories. We demonstrate the effectiveness of this approach on ulcerative colitis (UC) dataset. UC scoring remains a very challenging classification task, while being very important task for clinical decision making and minimising current subjectivity.
\begin{figure}[t!]
    \centering
    \includegraphics[scale=0.8]{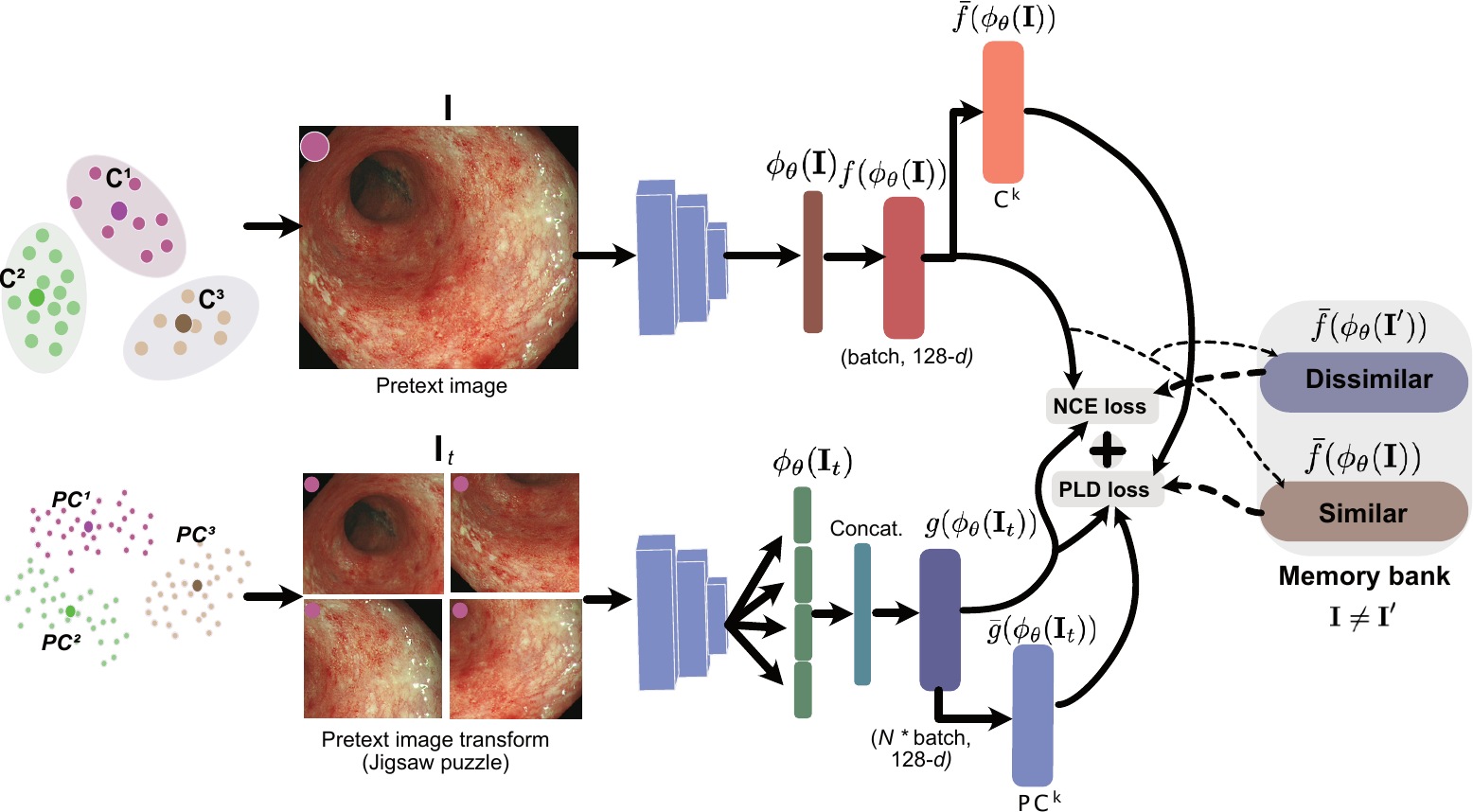}
    \caption{Pretext invariant patch-level instance group discrimination for ulcerative colitis (UC) scoring. Two identical classification networks are used to compute image-level and patch-level embedding. Three Mayo Endoscopic Scoring (MES) for UC from 1 up to 3 (mild: 1, moderate: 2 and severe: 3) are presented as three separate clusters for both images and patches. The memory bank contains the moving average of representations for all images in the dataset. Here, $\textbf{I}$ represent an image sample while $\textbf{I}_t$ is a transformed puzzle of that image and $\textbf{I}'$ represent negative sample.}
    \label{fig:blockDiagram}
\end{figure}
\section{Method}{\label{section:method}}
We propose to increase inter-class separation and minimise the intra-class distance by jointly minimising two loss functions that are based on contrastive loss. In contrast to widely used image similarity comparisons we use patch-level and image-level configurations. Additionally, we propose a novel instance group level discriminative loss. A memory bank is used to store moving average embedding of negative samples for efficient memory management. The block diagram of our proposed MES-scoring for ulcerative colitis classification framework is shown in Figure~\ref{fig:blockDiagram}.
\subsection{Pretext invariant representations}
Let the ulcerative colitis dataset consists of $N$ image samples denoted as $\mathcal{D}_{uc}=\{\textbf{I}_1, \textbf{I}_2, ..., \textbf{I}_N\}$ for which a transformation $\mathcal{T}$ is applied to create and reshuffle $m$ number of image patches for each image in $\mathcal{D}_{uc}$,  $\mathcal{P}_{uc}=\{\textbf{I}_{1t}^{1}, ...,\textbf{I}_{1t}^{m}, ..., \textbf{I}_{Nt}^{1}, ..., \textbf{I}_{Nt}^{m}\}$ with $\mathcal{T}\in t$. We train a convolutional neural network with free parameters $\theta$ that embody representation $\phi_\theta(\textbf{I})$ for a given sample $\textbf{I}$ and $\phi_\theta(\textbf{I}_t)$ for patch $\textbf{I}_t$. For image patches, representations of each patch constituting the image $\textbf{I}$ is concatenated. A unique projection heads, $f(.)$ and $g(.)$, are applied to re-scale the representations to a $128$-dimensional feature vector in each case (see Figure~\ref{fig:blockDiagram}). A memory bank is used to store positive and negative sample embedding of a mini-batch ${B}$ (in our case, $B = 32$). Negative refer to embedding for $I' \neq I$ that is required to compute our contrastive loss function $\mathcal{L}(.,.)$, measuring the similarity between two representations. The list of negative samples, say $\mathcal{D}_n$, grows with the training epochs and are stored in a memory bank $\mathcal{M}$. To compute a noise contrastive estimator (NCE), each positive samples has $|\mathcal{D}_n|$ negative samples and minimizes the loss:
\vspace{-0.25mm}
\small{
\begin{equation}{\label{eq:nce}}
   \mathcal{L}_{NCE}(\textbf{I},\textbf{I}_t) = -\log [h(f(\phi_\theta(\textbf{I})), g(\phi_\theta(\textbf{I}_t)))] - \sum_{\textbf{I}'\in \mathcal{D}_n} \log [1- h(f(\phi_\theta(\textbf{I}')), g(\phi_\theta(\textbf{I}_t)))].
\end{equation}}
\vspace{-0.25mm}
\noindent{For} our experiments we have used both ResNet50~\cite{he2016deep} and a combination of convolutional block attention (CBAM, ~\cite{woo2018cbam}) with ResNet50 model (ResNet50$^{\tiny{+cbam}}$) for computing the representation $f(.)$ and $g(.)$. In Eq.~(\ref{eq:nce}), $h(.,.)$ is the $cosine$ similarity between the representations with a temperature parameter $\tau$, and for $h(f(.), g(.))$:
\begin{equation}{\label{eq:cosine}}
    h(f, g) = \frac{\exp{<f, g>}/\tau}{\exp{<f, g>}/\tau + |\mathcal{D}_n|/N}.
\end{equation}

The presented loss encourages the representation of image $\textbf{I}$ to be similar to its corresponding transformed patches $\textbf{I}_t$ while increasing the distance between the dissimilar image samples $\textbf{I}'$. This enables network to learn invariant representations. 
\vspace{-0.25mm}
\subsection{Patch-level instance group discrimination loss}
Let $\bar{f}(.)$ and $\bar{g}(.)$ be the mean embedding for classes $k$ with cluster centers $C^{k}$ and $PC^{k}$ respectively for the image $\textbf{I}$ and patch samples $\textbf{I}_t$. $k$-means clustering is used to group the embedding into $k$ $(=3)$ class instances. The idea is to then compute the similarity of each patch embedding $g(.)$ with the mean image embedding $\bar{f}(.)$ for all $k$ classes and vice-versa using Eq.~(\ref{eq:cosine}). A cross-entropy (CE) loss is then computed that represent our proposed $L_{PLD}(.,.)$ loss function given as: 
\vspace{-1.5mm}
\small{
\begin{equation}{\label{eq:PLD_loss}}
\begin{split}
\mathcal{L}_{PLD}(\textbf{I},\textbf{I}_t) = -0.5\sum_{\forall k} C^k \log(h(\bar{f}(\phi_\theta(\textbf{I})), g(\phi_\theta(I_t))) \\ - 0.5
\sum_{\forall k} PC^k \log(h(\bar{g}(\phi_\theta(\textbf{I}_t)), f(\phi_\theta(\textbf{I}))).
\end{split}
\end{equation}
}
\noindent{The} proposed patch-level group discrimination loss $\mathcal{L}_{PLD}(\textbf{I},\textbf{I}_t)$ takes $k$ class instances into account. As a result not only the similarity between the group (mean) embedding and a single sample embedding for the same class is maximised but also it guarantees inter-class separation. The final loss function with empirically set $\lambda=0.5$ is minimised in our proposed PLD-PIRL network and is given by:
\vspace{-1.5mm}
\begin{equation}{\label{eq:final}}
    \mathcal{L}_{final}(\textbf{I}, \textbf{I}_t) = \mathcal{L}_{NCE}(\textbf{I}, \textbf{I}_t) + \lambda \mathcal{L}_{PLD}(\textbf{I}, \textbf{I}_t).
\end{equation}
\section{Experiments and results}
\subsection{Implementation Details.}
For training of pretext tasks in self-supervised learning, we use a learning rate (lr) of 1$e^{-3}$ and a SGD optimizer. 3000 epochs with a batch size of 32 is used to train pretext tasks presented in all experiments. For PLD loss, we set $k=3$ for number of clusters and test the effect of different temperature $\tau$ and $\lambda$ on the model performance. 

For the next downstream classification task, we use finetune the model with the learning rate of 1$e^{-4}$, the SGD optimizer, batch size of 32, and the learning rate decay strategy with a learning rate decay of 0.9 times per 30 epochs. Our experimental results showed that most of the models converged around 150 epochs. For the baseline supervised models training converged to higher epochs of nearly 200 epochs. The stopping criteria was based on minimal loss improvement of 0.000001 over 20 consecutive epochs. The proposed method is implemented on a server deployed with an NVIDIA Quadro RTX 6000 graphics card using the PyTorch framework. All input images were resized to $224\times224$ pixels.
%
\begin{table}[t!]
\centering
\caption{Experimental results on hold-out test set (Test-I) and unseen center test set (Test-II) for a three way classification of MES scores (1, 2 and 3). Both top 1 and top 2 accuracies are provided together with F1-score, specificity and sensitivity values (in percentage). Two best results are shown in bold. Both classical supervised approach (baseline) and different self-supervised methods are compared with our proposed PLD-PIRL classification method. \label{tab:UC-classification}}
\begin{tabular}{ll|c|c|c|c|c}
\hline
\multicolumn{1}{l|}{\textbf{Method}}                    & \multicolumn{1}{c|}{\textbf{Model}} & \multicolumn{1}{c|}{\textbf{Top 1}}  & \multicolumn{1}{c|}{\textbf{Top 2}}  & \multicolumn{1}{c|}{\textbf{F1}} & \multicolumn{1}{c|}{\textbf{Sen.}} & \textbf{Spec.} \\ 
\hline
\multicolumn{7}{l}{\textbf{Test-I}} \\ 
\hline
\multicolumn{1}{l|}{\multirow{2}{*}{\textbf{Baseline}}} & \multicolumn{1}{l|}{ResNet50}          & \multicolumn{1}{c|}{64.29\%}          & \multicolumn{1}{c|}{91.63\%}          & \multicolumn{1}{c|}{62.77\%}            & \multicolumn{1}{c|}{60.56\%}               & 81.02\%        \\  
\cline{2-7} 
    & \multicolumn{1}{|l|}{\begin{tabular}{@{}l@{}}ResNet50$^{\tiny{+cbam}}$ \end{tabular}}     & \multicolumn{1}{c}{65.47\%}         & \multicolumn{1}{c}{93.59\%}          & \multicolumn{1}{c}{64.38\%}            & \multicolumn{1}{c}{63.58\%}               & \multicolumn{1}{c}{81.95\%}        \\ 
   \hline
\multicolumn{1}{l|}{SimCLR~\cite{chen2020simple}}            & \multicolumn{1}{l|}{ResNet50}          & \multicolumn{1}{c|}{61.91\%}          & \multicolumn{1}{c|}{94.93\%}          & \multicolumn{1}{c|}{59.87\%}            & \multicolumn{1}{c|}{58.49\%}               & 79.60\%               \\ \hline
\multicolumn{1}{l|}{SimCLR~\cite{chen2020simple}}                               & {\begin{tabular}[t]{@{}l@{}}ResNet50$^{\tiny{+cbam}}$ \end{tabular}}       & \multicolumn{1}{c}{63.09\%}        & \multicolumn{1}{c}{92.27\%}        & \multicolumn{1}{c}{60.90\%}            & \multicolumn{1}{c}{59.31\%}             & \multicolumn{1}{c}{79.84\%}          \\ \hline
\multicolumn{1}{l|}{SimCLR + DCL~\cite{chuang2020debiased}}                       & \multicolumn{1}{l|}{ResNet50}          & \multicolumn{1}{c|}{64.28\%}          & \multicolumn{1}{c|}{\textbf{94.98\%}}         & \multicolumn{1}{c|}{62.34\%}            & \multicolumn{1}{c|}{62.78\%}               & 80.35\%               \\ 
\hline
\multicolumn{1}{l|}{SimCLR + DCL~\cite{chuang2020debiased}}      & ResNet50$^{+cbam}$ & \multicolumn{1}{c}{64.79\%} &\multicolumn{1}{c}{94.06\%}        &\multicolumn{1}{c}{62.01\%}           & \multicolumn{1}{c}{62.39\%} & \multicolumn{1}{c}{79.56\%}               \\ \hline
\multicolumn{1}{l|}{MOCO + CLD~\cite{he2020momentum}}        & \multicolumn{1}{l|}{ResNet50}          & \multicolumn{1}{c|}{66.96\%}          & \multicolumn{1}{c|}{93.12\%}          & \multicolumn{1}{c|}{65.79\%}            & \multicolumn{1}{c|}{66.38\%}               & 84.26\%               \\ \hline

\multicolumn{1}{l|}{MOCO + CLD~\cite{he2020momentum}}                               & ResNet50$^{+cbam}$       & \multicolumn{1}{c}{67.32\%}        & \multicolumn{1}{c}{92.52\%}        & \multicolumn{1}{c}{66.38\%}            & \multicolumn{1}{c}{65.91\%}            & \multicolumn{1}{c}{\textbf{84.53\%}}               \\ \hline
\multicolumn{1}{l|}{PIRL~\cite{misra2020self}}               & \multicolumn{1}{l|}{ResNet50}          & \multicolumn{1}{c|}{65.93\%}          & \multicolumn{1}{c|}{92.89\%}          & \multicolumn{1}{c|}{64.87\%}            & \multicolumn{1}{c|}{64.26\%}               & 81.03\%               \\ \hline
\multicolumn{1}{l|}{PIRL~\cite{misra2020self}}                               & ResNet50$^{+cbam}$       & \multicolumn{1}{c}{66.67\%}         & \multicolumn{1}{c}{93.21\%}        & \multicolumn{1}{c}{65.91\%}            & \multicolumn{1}{c}{66.47\%}             & \multicolumn{1}{c}{82.59\%}               \\ 
\hline
\multicolumn{1}{l|}{PLD-PIRL (ours)}  & \multicolumn{1}{l}{ResNet50}         &\multicolumn{1}{c}{\textbf{67.85\%}}          & \multicolumn{1}{c}{93.98\%}          & \multicolumn{1}{c}{\textbf{67.48\%}}          & \multicolumn{1}{c}{\textbf{66.93\%}}               & \multicolumn{1}{c}{83.41\%}             \\

\hline

\multicolumn{1}{l|}{PLD-PIRL(ours)}                        & ResNet50$^{+cbam}$     & \multicolumn{1}{c}{\textbf{69.04\%}} & \multicolumn{1}{c}{\textbf{96.31\%}} & \multicolumn{1}{c}{\textbf{68.98\%}}   & \multicolumn{1}{c}{\textbf{67.35\%}}      & \multicolumn{1}{c}{\textbf{84.71}\%}      \\ \hline

\multicolumn{7}{l}{\textbf{Test-II}}                                                \\ \hline
\multicolumn{1}{l|}{\multirow{2}{*}{\textbf{Baseline}}} & \multicolumn{1}{l|}{ResNet50}          & \multicolumn{1}{c|}{57.61\%}          & \multicolumn{1}{c|}{88.36\%}          & \multicolumn{1}{c|}{57.03\%}            & \multicolumn{1}{c|}{56.69\%}               & 71.10\%               \\ \cline{2-7} 
\multicolumn{1}{l|}{}                                   & ResNet50$^{+cbam}$     &\multicolumn{1}{c}{60.92\%} & \multicolumn{1}{c}{90.49\%}          & \multicolumn{1}{c}{59.38\%}            &\multicolumn{1}{c}{58.81\%}            & \multicolumn{1}{c}{73.79\%}               \\ \hline
\multicolumn{1}{l|}{SimCLR~\cite{chen2020simple}}                             & \multicolumn{1}{l|}{ResNet50}          & \multicolumn{1}{c|}{56.95\%}          & \multicolumn{1}{c|}{85.88\%}          & \multicolumn{1}{c|}{55.91\%}            & \multicolumn{1}{c|}{54.69\%}               & 71.26\%               \\ \hline
\multicolumn{1}{l|}{SimCLR~\cite{chen2020simple}}                               & ResNet50$^{+cbam}$      & \multicolumn{1}{c}{57.31\%}         & \multicolumn{1}{c}{85.21\%}          & \multicolumn{1}{c}{56.29\%}            & \multicolumn{1}{c}{56.50\%}              & \multicolumn{1}{c}{71.98\%}               \\ \hline
\multicolumn{1}{l|}{SimCLR + DCL~\cite{chuang2020debiased}}                       & \multicolumn{1}{l|}{ResNet50}          & \multicolumn{1}{c|}{58.94\%}          & \multicolumn{1}{c|}{87.92\%}          & \multicolumn{1}{c|}{57.36\%}            & \multicolumn{1}{c|}{57.29\%}               & 73.29\%               \\ \hline
\multicolumn{1}{l|}{SimCLR + DCL~\cite{chuang2020debiased}}                               & ResNet50$^{+cbam}$     & \multicolumn{1}{c}{59.60\%}         & \multicolumn{1}{c}{90.34\%}        & \multicolumn{1}{c}{58.42\%}            & \multicolumn{1}{c}{59.58\%}             & \multicolumn{1}{c}{74.19\%}               \\ \hline
\multicolumn{1}{l|}{MOCO + CLD~\cite{he2020momentum}}                         & \multicolumn{1}{l|}{ResNet50}          & \multicolumn{1}{c|}{60.61\%}          & \multicolumn{1}{c|}{90.71\%}          & \multicolumn{1}{c|}{60.52\%}            & \multicolumn{1}{c|}{59.88\%}               & 75.69\%               \\ \hline
\multicolumn{1}{l|}{MOCO + CLD~\cite{he2020momentum}}                               & ResNet50$^{+cbam}$     & \multicolumn{1}{c}{60.93\%}          & \multicolumn{1}{c}{92.12\%}         & \multicolumn{1}{c}{60.61\%}           & \multicolumn{1}{c}{59.29\%}             & \multicolumn{1}{c}{77.33\%}               \\ \hline
\multicolumn{1}{l|}{PIRL~\cite{misra2020self}}                               & \multicolumn{1}{l|}{ResNet50}          & \multicolumn{1}{c|}{61.59\%}          & \multicolumn{1}{c|}{92.61\%}          & \multicolumn{1}{c|}{60.55\%}            & \multicolumn{1}{c|}{60.53\%}               & 75.98\%               \\ \hline
\multicolumn{1}{l|}{PIRL~\cite{misra2020self}}                               & ResNet50$^{+cbam}$      &\multicolumn{1}{c}{62.25\%}          & \multicolumn{1}{c}{\textbf{93.92\%}}          & \multicolumn{1}{c}{61.96\%}            & \multicolumn{1}{c}{60.92\%}               & \multicolumn{1}{c}{78.03\%}               \\ \hline
\multicolumn{1}{l|}{PLD-PIRL (ours)}   & \multicolumn{1}{l}{ResNet50}          & \multicolumn{1}{c}{\textbf{62.90\%}}          & \multicolumn{1}{c}{92.93\%}          & \multicolumn{1}{c}{\textbf{62.81\%}}       & \multicolumn{1}{c}{\textbf{61.79\%}}               & \multicolumn{1}{c}{\textbf{80.23}\%}              \\ \hline
\multicolumn{1}{l|}{{\begin{tabular}[t]{@{}l@{}} PLD-PIRL(ours)\end{tabular}}}                        & ResNet50$^{+cbam}$   & \multicolumn{1}{c}{\textbf{64.24\%}} &\multicolumn{1}{c}{\textbf{95.32\%}} & \multicolumn{1}{c}{\textbf{64.38\%}}   & \multicolumn{1}{c}{\textbf{62.99\%}}     & \multicolumn{1}{c}{\textbf{80.09}\%}  \\ \hline
\multicolumn{7}{l}{\footnotesize{Sen. - sensitivity;} \hspace{.1cm} \footnotesize{Spec. - specificity}}
\end{tabular}
\end{table}
\subsection{Datasets and evaluation.}
We have used both publicly available and in-house dataset. HyperKvasir~\cite{borgli2020hyperkvasir} public dataset was used for model training, validation and as hold-out test samples (referred as Test-I). The available dataset includes MES scores (1,2 and 3) and three additional scoring levels categorising into scores 0-1, 1-2 and 2-3, totaling to 6 UC categories and 851 images. After re-examination by expert colonoscopist, the final data was divided into three different grades: mild, moderate and severe. In the experiment, 80\% of the data is used for training, 10\% is used for validation and 10\% is used for testing. Furthermore, to evaluate the efficacy of the proposed PLD-PIRL method on unseen center data (referred as Test-II), we used one in-house dataset as the test set. This dataset contains 151 images from 70 patient videos. We manually selected frames containing UC from the videos, which were then labeled as mild, moderate, and severe by an expert colonoscopist. All datasets will be made public upon acceptance of the paper.

We have used standard top-$k$ accuracy (percentage of samples predicted correctly), F1-score ($=\frac{tp}{tp+fp}$, tp: true positive, fp: false positive), specificity ($=\frac{tp}{tp+fn}$) and sensitivity ($=\frac{tn}{tn+fp}$) and for our 3-way classification task of MES-scoring for UC. 
\subsection{Comparison with SOTA methods}
Result of baseline fully supervised classification and self-supervised learning model (SSL) for UC classification on two test datasets (Test-I and Test-II) are presented in Table~\ref{tab:UC-classification}. ResNet50 and ResNet50$^{\tiny{+cbam}}$ are established as the baseline model for supervised learning and the same are also used for other state-of-the-art (SOTA) SSL comparisons. In Table~\ref{tab:UC-classification} for Test-I dataset, it can be observed that the proposed PLD-PIRL approach using ResNet50$^{\tiny{+cbam}}$ model achieves the best results with 69.04\%, 68.98\%, 84.71\% and 67.35\%, respectively, for top 1 accuracy, F1 score, specificity and sensitivity. Compared to the supervised learning based baseline models i.e., ResNet50 and ResNet50$^{\tiny{+cbam}}$, the top 1 accuracy is improved by 4.75\% and 3.57\%, respectively, for these models using our proposed PLD-PIRL. We also compared the proposed PLD-PIRL approach with other SOTA self-supervised learning methods including popular SimCLR~\cite{chen2020simple}, SimCLR+DCL~\cite{yeh2021decoupled}, MOCO+CLD~\cite{he2020momentum} and PIRL~\cite{misra2020self} methods. Our proposed network (ResNet50) clearly outperforms all these methods with at least nearly 1.1\% (MOCO+CLD) up to 6\% (SimCLR)  on top-1 accuracy. %

Similarly, for out-of-sample unseen center Test-II dataset (see Table~\ref{tab:UC-classification}), the proposed model outperforms the baseline fully supervised models by a large margin accounting to nearly 5\% for ResNet50 and 4\% for ResNet50$^{\tiny{+cbam}}$. A similar trend is observed for all SOTA SSL methods ranging from 3.63\% for MOCO+CLD upto 5.3\% for SimCLR with ResNet50. A clear boost of 1.99\% can be seen for the best PLD-PIRL (ResNet50$^{\tiny{+cbam}}$) model compared to the PIRL (ResNet50$^{\tiny{+cbam}}$).
Our experiments on all the existing approaches with ResNet and CBAM (ResNet$^{\tiny{+cbam}}$) backbone showed nearly 1\% improvement over ResNet50 on test-I with other SOTA (e.g., top 1 accuracies for SimCLR: 63.09\%, SimCLR+DCL: 64.79\% and for MOCO+CLD: 67.32\%) and around 0.5\% on test-II (SimCLR: 57.31\%, SimCLR+DCL: 59.60\% and for MOCO+CLD: 60.93\%).
Figure~\ref{fig:tsneplot} (left) representing $t$-SNE plots demonstrate an improved separation of sample points in both training and test sets compared to fully supervised baseline approach. Confusion matrix for both Test-I and Test-II are provided in supplementary Figure 1 that shows that proposed PLD-PIRL were able to classify more samples compared to the baseline method. Similarly, it can be observed from supplementary Figure 2 that the wrongly classified ones are only between adjacent classes which at times categories as both classes by the clinical endoscopists.
\subsection{Ablation study}
\begin{figure}[t!b!]
    \centering
    \includegraphics[width=\textwidth]{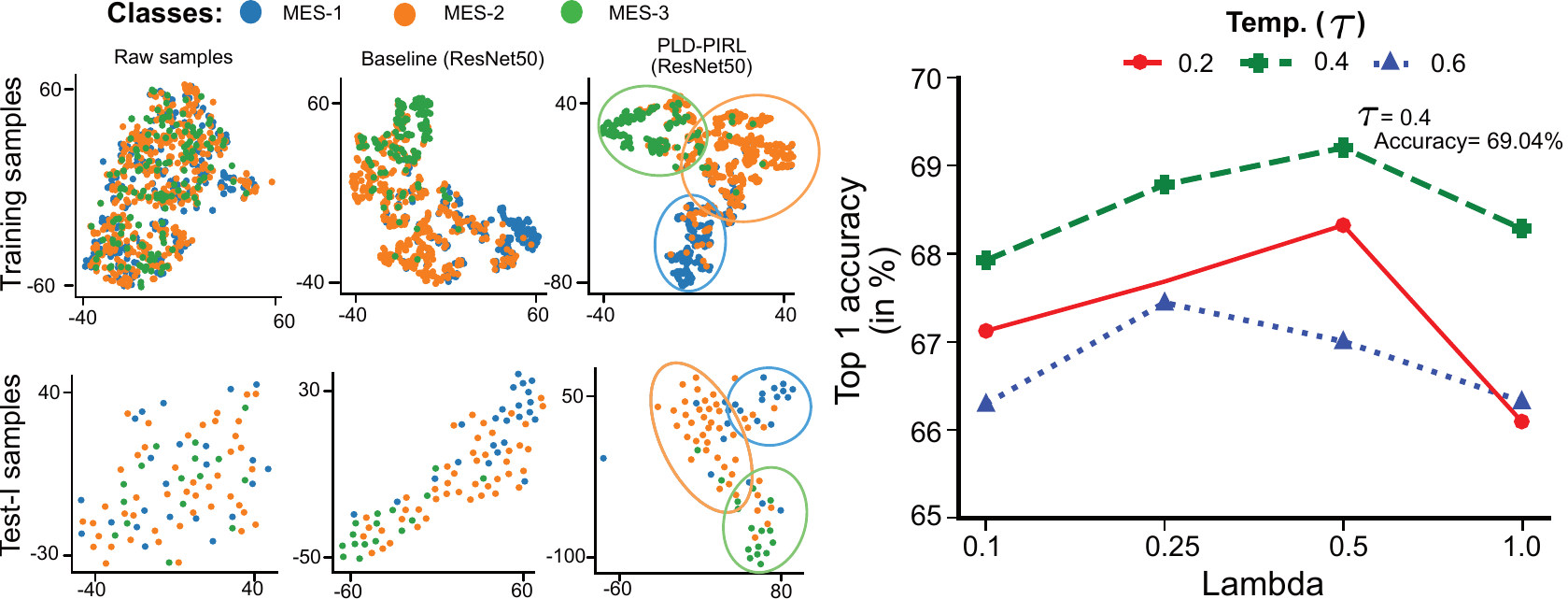}
    \caption{(Left) Classified clusters for three MES classes obtained from fully supervised baseline and proposed PLD-PIRL on both training (top) and test-I (down) samples. Raw sample distributions are also shown. A $t$-distributed stochastic neighbor embedding is used for the point plots of image samples embedding. (Right) Experiments for finding best values for hyper-parameters temperature $\tau$ and $\lambda$ weights in the loss function.}
    \label{fig:tsneplot}
\end{figure}

Our experiments indicate that the settings of $\lambda$ and temperature $\tau$ parameters in the proposed PLD-PIRL approach will affect the model performance. Therefore, we conducted an ablation study experiment to further study the performance of PLD-PIRL under different parameter settings. We set $\tau = \{0.2, 0.4, 0.6\}$ and $\lambda=\{0.1, 0.25, 0.5, 1.0\}$. As can it can be observed from the plot in Figure~\ref{fig:tsneplot} (right) that for  $\tau =0.4$ PLD-PIRL maintained high accuracy at different $\lambda$ values, and is better than other parameter settings. The best value is obtained at $\lambda=0.5$ with the top1 accuracy of 69.04\%.

%
\section{Conclusion}
Our novel self-supervised learning method using pretext-invariant representation learning with patch-level instance-group discrimination (PLD-PIRL) applied to the UC classification task overcomes the limitations of previous approaches that rely on binary classification tasks. We have validated our method on a public dataset and an unseen dataset. Our experiments show that compared with other SOTA classification methods that include fully supervised baseline models, our proposed method obtained large improvements in all metrics. The test results on the unseen dataset provides an evidence that our proposed PLD-PIRL method can learn to capture the subtle appearance of mucosal changes in colonic inflammation and the learnt feature representations together with instance-group discrimination allows improved accuracy and robustness for clinically use Mayo Endoscopic Scoring of UC. 
\bibliographystyle{splncs04}
\bibliography{main}
\section*{Supplementary material}
\begin{suppfigure}[h!]
    \centering
    \includegraphics[width=\textwidth]{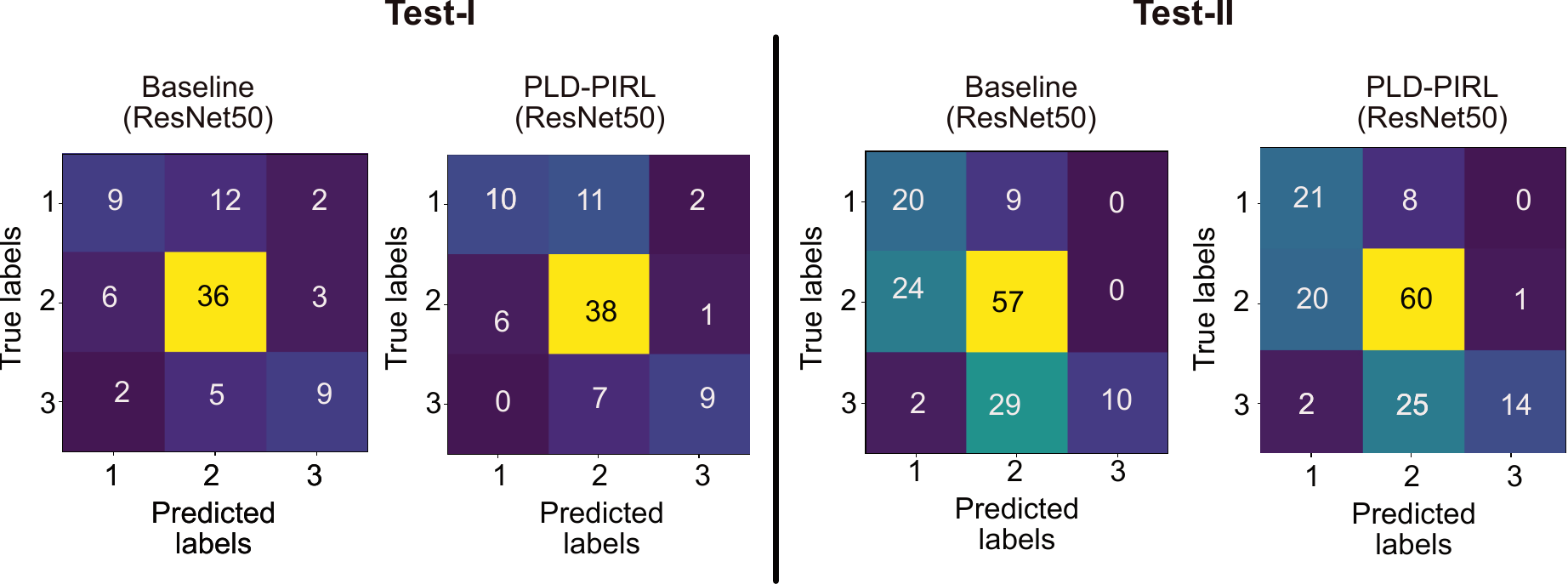}
    \caption{Confusion matrix for our proposed method (PLD-PIRL) and the baseline fully supervised method both using ResNet50 model on test-I dataset (same distribution, left from the solid line) and test-II dataset (out-of-sample distribution, right from the solid line).}
    \label{fig:TGANet}
\end{suppfigure}
\begin{suppfigure}[t!]
    \centering
    \includegraphics[width=\textwidth]{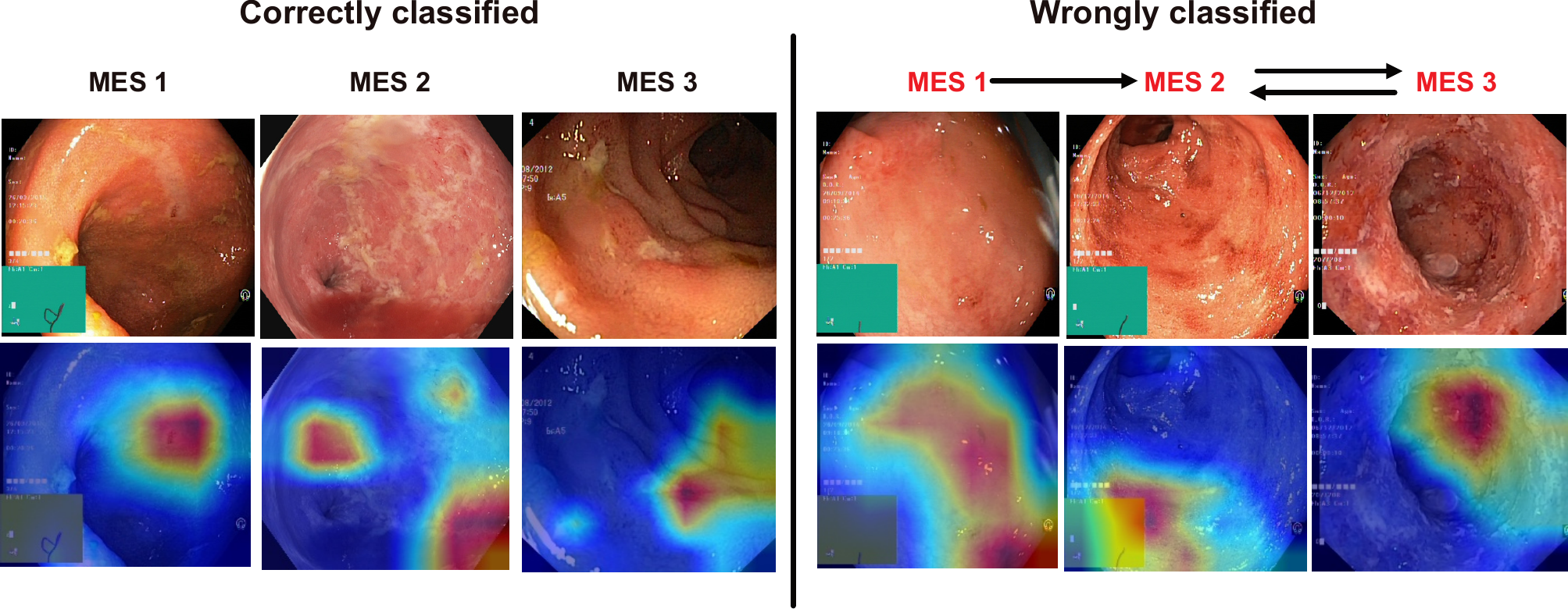}
    \caption{Qualitative results showing the model attentions for correctly (left from the solid line) and incorrectly (right from the solid line) classified Mayo Endoscopic Scores. Arrows in wrongly classified ones points to the predicted score for the given sample.}
    \label{fig:TGANet}
\end{suppfigure}

\end{document}